\documentclass[conference]{IEEEtran}
\IEEEoverridecommandlockouts
\usepackage{cite}

\usepackage[usenames,dvipsnames]{xcolor} 
\usepackage{hyperref} 
\definecolor{linkcolour}{rgb}{0,0.2,0.6} 
\hypersetup{colorlinks,breaklinks,urlcolor=linkcolour,linkcolor=linkcolour} 

\usepackage{amsmath,amssymb,amsfonts}
\usepackage{algorithmic}
\usepackage{siunitx}
\usepackage{mathtools}
\usepackage{graphicx}
\graphicspath{{pictures/}}
\usepackage{subfig}
\usepackage{textcomp}
\usepackage{xcolor}
\usepackage{etoolbox}
\usepackage{mathptmx}
\def\BibTeX{{\rm B\kern-.05em{\sc i\kern-.025em b}\kern-.08em
    T\kern-.1667em\lower.7ex\hbox{E}\kern-.125emX}}

\begin{document}

\title{Experimental Study on the Imitation of the Human Head-and-Eye Pose Using the 3-DOF Agile Eye Parallel Robot with ROS and Mediapipe Framework\\
}
\author{\IEEEauthorblockN{Amirmohammad Radmehr}
\IEEEauthorblockA{\textit{University of Tehran} \\
\textit{School of Electrical and}\\ \textit{Computer Engineering}\\
Tehran, Iran\\
amir.radmehr77@gmail.com}
\and
\IEEEauthorblockN{Milad Asgari}
\IEEEauthorblockA{\textit{University of Tehran} \\
\textit{School of Electrical and}\\ \textit{Computer Engineering}\\
Tehran, Iran \\
miladasgari178@gmail.com}
\and
\IEEEauthorblockN{Mehdi Tale Masouleh}
\IEEEauthorblockA{\textit{University of Tehran} \\
\textit{School of Electrical and}\\ \textit{Computer Engineering}\\
\href{https://taarlab.com/}{\textit{Human and Robot Interaction Laboratory }}\\
Tehran, Iran\\
m.t.masouleh@ut.ac.ir}
}

\maketitle

\begin{abstract}

In this paper, a method to mimic a human face and eyes is proposed which can be regarded as a combination of computer vision techniques and neural network concepts. From a mechanical standpoint, a 3-DOF spherical parallel robot is used which imitates the human head movement. In what concerns eye movement, a 2-DOF mechanism is attached to the end-effector of the 3-DOF spherical parallel mechanism. In order to have robust and reliable results for the imitation,  meaningful information should be extracted from the face mesh for obtaining the pose of a face, i.e., the roll, yaw, and pitch angles.
To this end, two methods are proposed where each of them has its own pros and cons. The first method consists in resorting to the so-called Mediapipe library which is a machine learning solution for high-fidelity body pose tracking, introduced by Google. As the second method, a model is trained by a linear regression model for a gathered dataset of face pictures in different poses. 
In addition, a 3-DOF Agile Eye parallel robot is utilized to show the ability of this robot to be used as a system which is similar to a human head for performing a 3-DOF rotational motion pattern. Furthermore, a 3D printed face and a 2-DOF eye mechanism are fabricated to display the whole system more stylish way. Experimental tests, which are done based on a ROS platform, demonstrate the effectiveness of the proposed methods for tracking the human head and eye movement.

\end{abstract}


\section{Introduction}

Human Robot Interaction (HRI) is one of the most important tasks in social robotics. In the last decades, HRI has become an interesting research where different untrained users interact with robots in real scenarios. In the last decade, effective HRI has become an exciting topic for research. Facial expressions are rich sources of information about effective behavior and have been commonly used for emotion recognition. In this respect, imitating a human face are one of the most critical elements in a natural HRI\cite{Breazeal}.
In the last two decades, essential tools such as image processing techniques have been developed, which enables computers to analyze images in a higher dimension\cite{Mao}. These tools allow researchers to analyze complex issues such as facial expressions or face pose. The issue of finding pose of a face has several challenges which can be solved with various methods. Finding a facial pose is a feature that, while trivial to humans, is a significant challenge for computers. From a mechanical stand point, the  latter movement can be made equiavelent to the motion exihibited by a 3-DOF spherical robot.

Today, neural networks are used more than ever. Furthermore, many studies have been done in this field, one of which is processing images and extracting information from them. One of the challenges which can be solved using neural network is finding angles of a face. One of the methods to obtain the angles of a human face is the nonlinear regression approach \cite{jones2003fast, rowley1998rotation}. In this method, a set of data is collected from different angles then a nonlinear function is fitted on them.

Another method to obtain these angles is using a pool of images with different poses  of several faces. So that each pose has a unique label and to get a label for a new pose, the similarity of that pose to the pool is checked \cite{ng2002composite}. However, these errors can be eliminated by using appropriate filters and various linear transformations, but these angles are still not accurately obtained \cite{sherrah2001face} \cite{wilson2000perception}.

An alternative approach to attain face angles is using SOLVEPNP\_P3P \cite{mallick_2021}, which is based on linear transformations and uses a 68-point detector to obtain several specific points on the face. In the latter approach, a three-dimensional perspective is created, which uses Efficient Perspective-N-Point (EPNP) camera pose estimation \cite{lepetit2009epnp} to solve the problem using these points. The main problem with this method is the workspace and its unreliability. During the tests performed with this method, the output fluctuated a lot with a slight face rotation.

\begin{figure*}[t]
\centerline{\includegraphics[width=2\columnwidth]{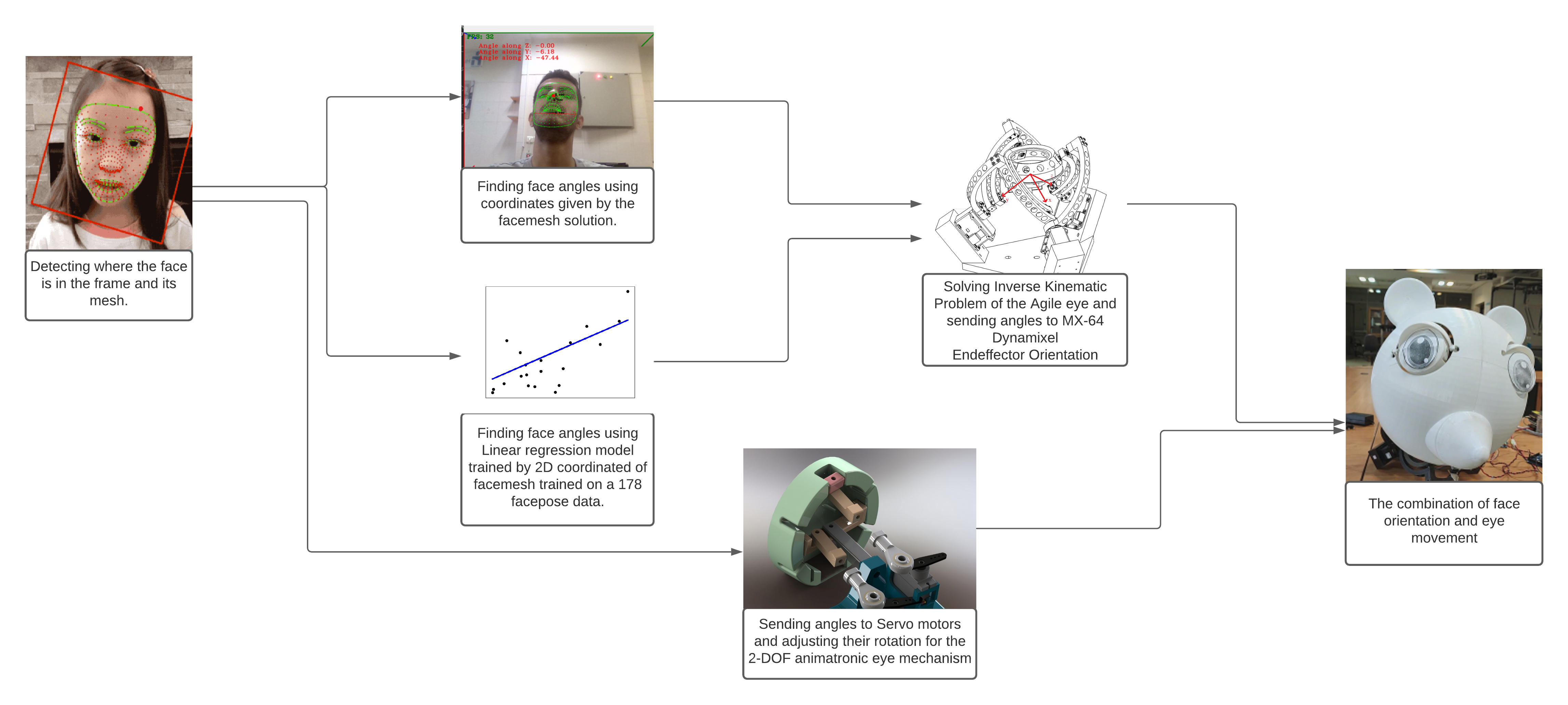}}
\caption{Overview of the overall system.}
\label{flowchart}
\end{figure*}

However, one can use the MediaPipe platform \cite{lugaresi2019mediapipe} to find angles of a face. This platform allows having a three-dimensional mesh of the face inside the image. This high-speed platform (about 30 frames per second) can be used for real-time application. This library is used to find face angles using two methods elaborated in Sections II and III.

Many mechanical systems have been proposed for mimicking the human movements which most of them are based on serial structures and a few of them are based on parallel ones ~\cite{ceccarelli2020parallel}.
In the preceding decades, several parallel robots have been developed to mimic human movements, such as leg or arm imitation \cite{serialleg} \cite{armrob}.
Parallel robots can be used to control the movement of the human head; one can use a 2-DOF mechanism \cite{neck2dof} or also a serial mechanism \cite{neck2dof2}.
Although there exists many parallel robotic platform which can mimic the human head motion, most of them have only two DOF, with the yaw motion being actuated separately, and also some works have been done on 3-DOF parallel robots to mimic human head movements\cite{Gao},\cite{Lingampally}. The Agile Eye robot with 3-DOF \cite{gosselin1996development}, due to its high accuracy and speed, is an applicable option to imitate head movement.

The main contribution of this study consists in detecting the face angles using the combination of Mediapipe network and linear regression model. Moreover, these methods are applied to the Agile Eye parallel robot in real-time manner with high accuracy for imitating head and eye movement. In addition, it has been also shown that the Agile Eye 3-DOF parallel robot can be used as the head of a humanoid robot.

The remainder of this paper is organized as follows. In Section II, the system components are formulated in order to clarify how the system works. Then, the inverse and forward kinematics of the 3-DOF Agile Eye robot are investigated to apply face angles calculated by the proposed methods to this robot. Then a PID controller and motion planning are explored to achieve the best velocity and acceleration for three motors to reach the desired angles. Section III deals with methods to find facial angles by two ways to capture a human face angles. Finally, in Section IV, the paper concludes with some hints and remarks as ongoing works.

\begin{figure}[t]
\centerline{\includegraphics[width=75mm,scale=0.5]{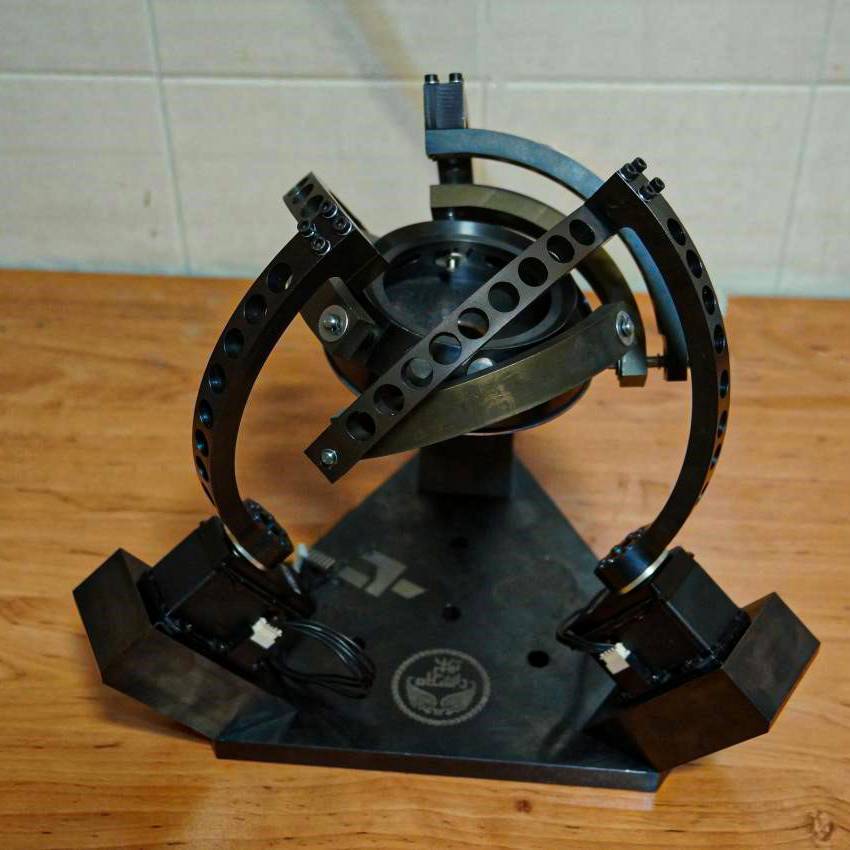}}
\caption{3-DOF Agile eye parallel robot manufactured at the Human and Robot Interaction Laboratory, University of Tehran which was first introduced in \cite{gosselin1996development}.}
\label{agileeye}
\end{figure}



\section{System Components}

This section demonstrates all components of the whole system and discusses the details of each component. Firstly, the 3-DOF Agile Eye parallel robot is explained, acting as the system's head. Then a 2-DOF eye mechanism of the robot and fabrication of a head is presented, and finally, methods regarding finding face orientation are described.

As aforementioned, the objective of the fabricated system consists in tracking a face in real-time. The face angles are calculated; after that, these angles will be sent to the Agile Eye parallel robot to move in the desired position. Simultaneously, the position of the face in the frame is tracked, and proper commands are sent to the 2-DOF eye mechanism attached to the end-effector of the 3-DOF Agile Eye parallel robot.


\subsection{3-DOF Agile Eye Parallel Robot}

Degrees-Of-Freedom (DOF) is a crucial concept in choosing a mechanism which imitates actions similar to the human head. It may be evident that human head poses can be summarized in three DOFs; roll, pitch, and yaw. The 3-DOF Agile Eye parallel robot which is shown in Fig.~\ref{agileeye} is a suitable choice to imitate the desired actions.

\begin{figure}[b]
\centerline{\includegraphics[width=75mm,scale=0.5]{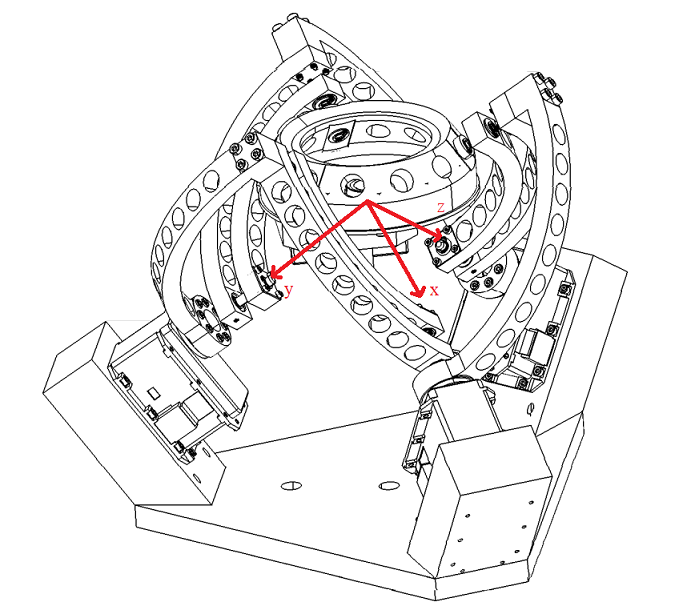}}
\caption{The \textit{Agile Eye} at its reference configuration.}
\label{agileeye_ref}
\end{figure}

In order to imitate the human head, it is needed to know the angles of the end-effector; this actively means that inverse kinematics of the Agile Eye parallel robot should be determined. This study addresses both forward kinematics and inverse kinematics of the robot\cite{ikp}. First the Inverse Kinematic Problem (IKP) of the robot is elaborated. According to the following relations 
having  $\phi$ (along $z$), $\theta$ (along $y$)and $\psi$ (along $x$) leads to obtain the exact rotation for three base joints relative to its reference configuration which is shown in Fig.~\ref{agileeye_ref}:
\\\\

\begin{equation}
\tan\theta_1 = \frac{\cos\theta \sin\psi}{\cos\phi \cos\psi + \sin\phi \sin\theta \sin\psi}\label{ikp1}
\end{equation}
\begin{equation}
\tan\theta_2 = \frac{\sin\phi \sin\psi + \cos\phi \sin\theta \cos\psi}{\cos\theta \cos\psi}\label{ikp2}
\end{equation}
\begin{equation}
\tan\theta_3 = \tan\phi \label{ikp3}
\end{equation}

\begin{figure}[t]
\centering
\subfloat[Rotation axes are shown as red vectors.]{\includegraphics[width=0.5\columnwidth]{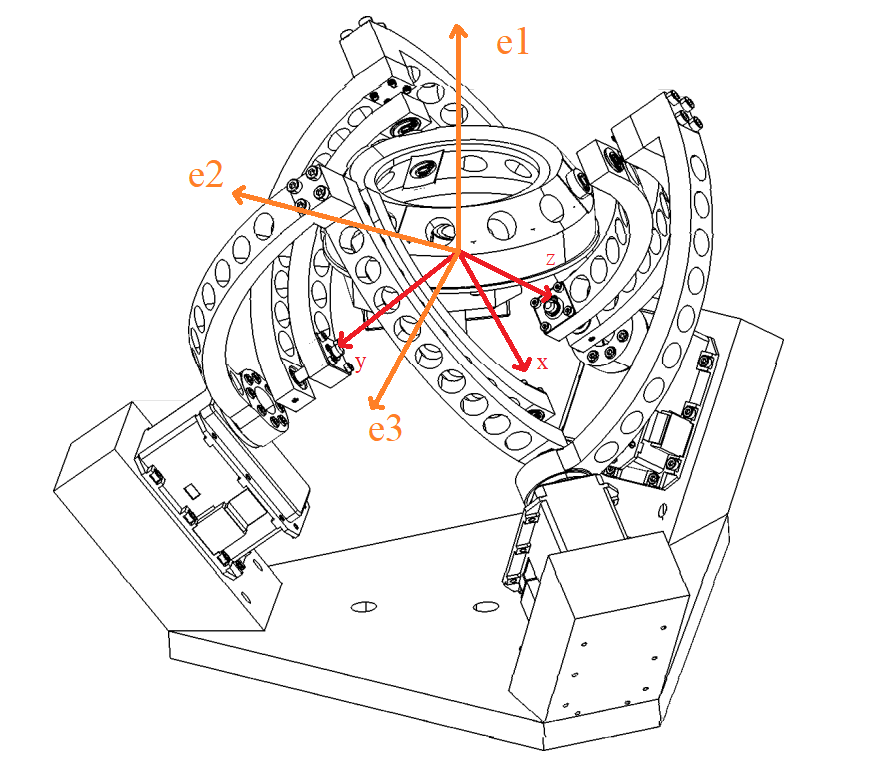}}
\hfil
\subfloat[Three degree of freedom of human head.]{\includegraphics[width=0.5\columnwidth]{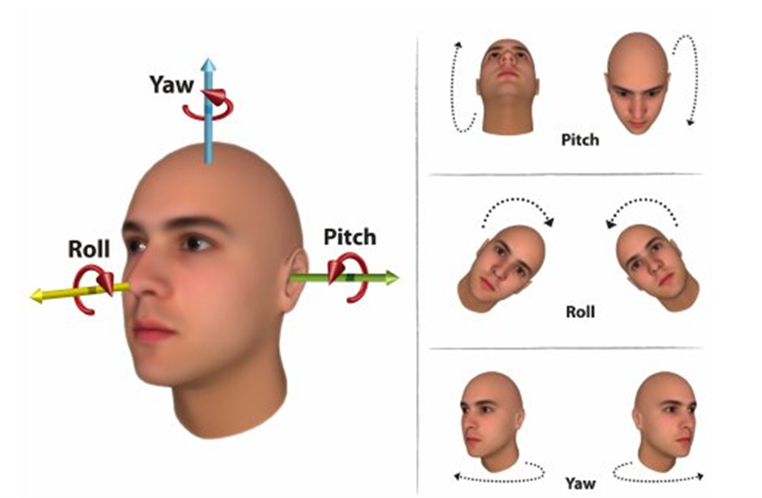}}
\caption{Configuring axes of rotations to match human head rotation axes.}
\label{newaxes}
\end{figure}

In order to use the Agile Eye as an imitation of head movement, new rotation axes should be proposed; these axes correspond to the rotation of the human head, as shown in Fig.~\ref{newaxes}. $e_1$ as yaw axis, $e_2$ as roll axis and $e_3$ as pitch axis. The following relation demonstrates rotation matrices corresponding to each mentioned vectors:

\begin{equation}
\vec{\mathbf{e_1}}= \frac{1}{\sqrt{3}} \begin{bmatrix} -1  \\ -1 \\ -1 \end{bmatrix}
\vec{\mathbf{e_2}} = \frac{\sqrt{2}}{\sqrt{3}}\begin{bmatrix} -1  \\ 0.5 \\ 0.5 \end{bmatrix}
\vec{\mathbf{e_3}} = \frac{\sqrt{2}}{3}\begin{bmatrix} 0  \\ 1.5 \\ -1.5 \end{bmatrix}
\label{vectors}
\end{equation}

\begin{equation}
\mathbf{Q_i} = \vec{\mathbf{e_i}}\vec{\mathbf{e_i}}^T + (I - \vec{\mathbf{e_i}}\vec{\mathbf{e_i}}^T)\cos a_i + \mathbf{CPM}(\vec{\mathbf{e_i}}) \sin a_i
\label{rotQ}
\end{equation}
\begin{equation}
\mathbf{Q} = \mathbf{Q_x} \mathbf{Q_y} \mathbf{Q_z}
\end{equation}
\begin{equation}
\mathbf{CPM}(\vec{\mathbf{e_i}}) = \begin{bmatrix}
 0 & -{e_i}_z & {e_i}_y  \\ 
 {e_i}_z & 0 & -{e_i}_x \\ 
-{e_i}_y & {e_i}_x & 0
\end{bmatrix}
\label{CPM}
\end{equation}

\subsection{Control}

The controller which is considered for this project is a simple PID controller\cite{pid}. The signal which is controlled is the angular position of each servo motor. Since data are sent out continuously to the servos; a controller should be utilized to adjust the servos' motion in order to have a smooth maneuver for the end-effector of the Agile Eye. It should be noted that the P, I, and D parameters are tuned by performing several  trial and error tests.

\begin{figure}[t]
\centerline{\includegraphics[width=\columnwidth]{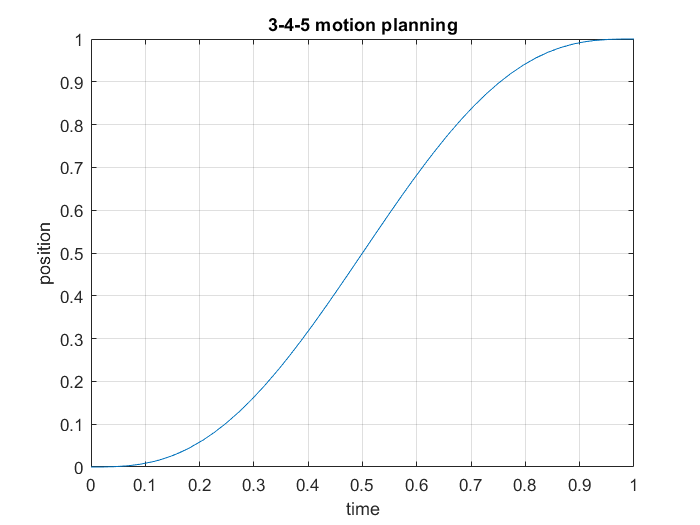}}
\caption{A 3-4-5 motion planning trajectory.}
\label{345_pic}
\end{figure}

Another subject which should be considered is the trajectory between every two positions assigned to the motors. As aforementioned, the Agile Eye has to imitate head movement, and to do so, between every two positions that settle, the acceleration at the beginning of the movement and at the end should be zero. In this application, a 3-4-5 motion planning operation \cite{angeles2002fundamentals} based on the following relation seems proper:

\begin{equation}
S(0) = 0, \frac{dS}{d\tau}|_{\tau=0}= 0
\end{equation}
\begin{equation}
S(1) = 1, \frac{dS}{d\tau}|_{\tau=1}= 0
\end{equation}
\begin{equation}
S(\tau) = 6\tau^5 - 15\tau^4 + 10\tau^3
\label{345}
\end{equation}
As shown in Fig.~\ref{345_pic}, at the start of a motion, position starts to change at a prolonged rate (at point zero with 0 velocities) and then increases the speed. At the midpoint of the motion, it starts to ease the acceleration and stops at the track's end.


\subsection{Robot Operating System}

In order to make the connection between both the image processing system that operates in real-time and the mechanical system, the Agile Eye robot, consideration of an operating system is necessary. Since mechanical systems are much slower than computer systems, a platform like ROS should be employed to execute processes simultaneously. Other options like threading or multiprocessing are available, but ROS is more reliable and easier to debug.

\begin{figure}[b]
\centerline{\includegraphics[width=\columnwidth]{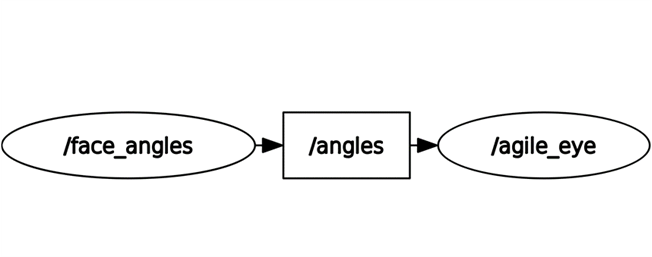}}
\caption{ROS computation graph containing nodes and topics.}
\label{ros_pic}
\end{figure}

As it is shown in Fig.~\ref{ros_pic}, there are two nodes, Node ``face\_angles" and the node ``agile\_eye", and one topic; ``angles''.
\subsubsection{face angles}
Node ``face\_angles" is the image processing component where three angles of a face are calculated and then published to ``angles" topic. 
\subsubsection{angles}
Topic ``angles'' is the data that a subscriber in the system uses. When the publisher gathers data, those data are prepared for any subscriber in the system to use for any purpose it fits. In this case, the agile\_eye node is the subscriber.
\subsubsection{agile eye}
Node ``agile\_eye" is a control system for the Agile Eye where inverse kinematic of the robot is considered. Based on the end-effector angles which are subscribed from the topic ``angles", proper positions are assigned to the three servos which are MX-106 Dynamixel servo motors. . In conclusion, the script responsible for this part contains the IKP of the robot, the PID controller, and the path planning generator function.


\subsection{2-DOF Eye Mechanism}

\begin{figure}[htbp]
\centering
\subfloat[Vertical axes.]{\includegraphics[width=0.5\columnwidth]{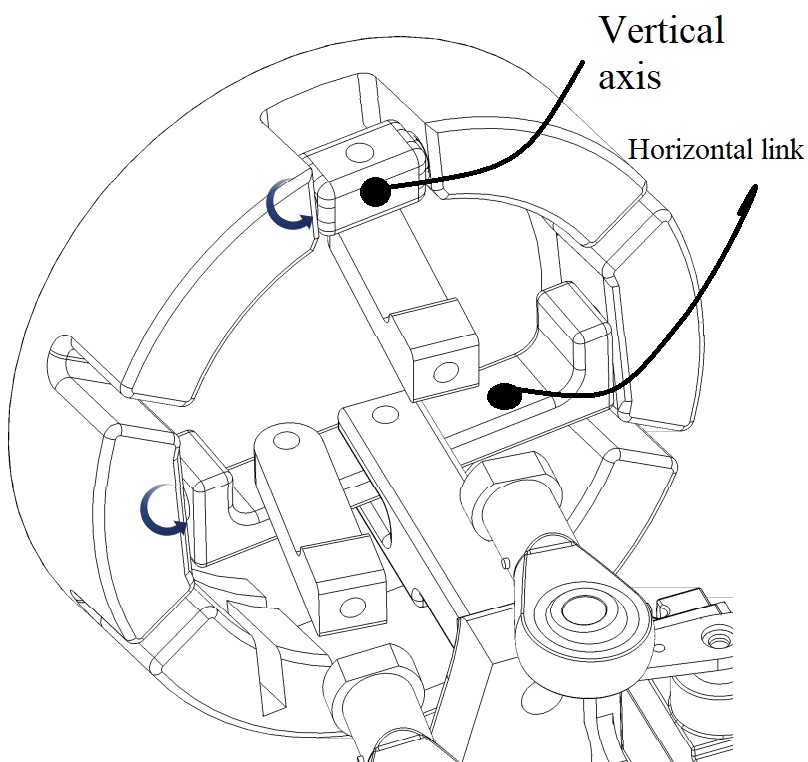}}
\hfil
\subfloat[Horizontal axes.]{\includegraphics[width=0.5\columnwidth]{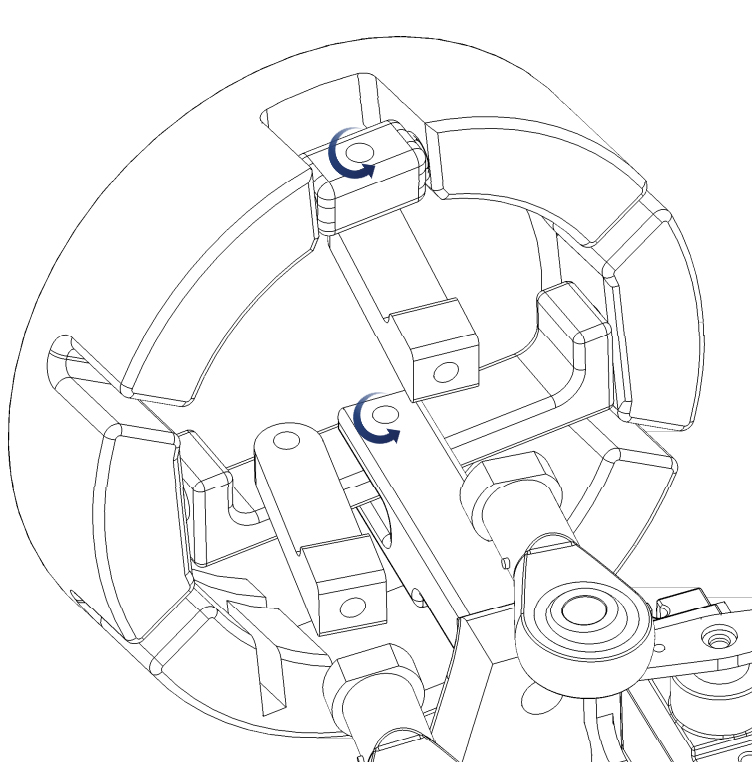}}
\caption{Axes which make the eyeball rotate freely both vertically and horizontally.}
\label{eye2}
\end{figure}

In this part the is a 2-DOF animatronic eye mechanism is introduced which increase the overall system's liveliness. The 2-DOF eye mechanism has a chassis which holds the horizontal link, and at its center, there is a joint that the whole eye revolves around, as shown in Fig.~\ref{eye2}. Another link shown in the latter figure is attached to the horizontal link and a servo in the back which moves the whole eye horizontally. 

The link and joint shown in Fig.~\ref{eye2} are responsible to control the eye vertically. This joint is also attached to a link connected to another servo to act vertically. One of the DOF is shown in Fig.~\ref{eye2}(a), and the other is the angle shown in Fig.~\ref{eye2}"(b). Notice that these two actions are separate from each other and do not affect each other. The condition for this circumstance is that the vertical and horizontal axes have to be in the same plane in 3D space in the reference configuration. 



%


\subsection{Facepose Estimation}

\begin{figure}[htbp]
\centerline{\includegraphics[width=\columnwidth]{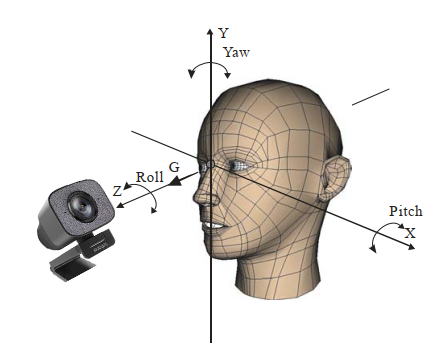}}
\caption{Animatronic eye horizontal and vertical axes.}
\label{cam_pic}
\end{figure}

As aforementioned, calculating the orientation of a human face in front of the camera is a challenging problem. After deploying the image processing tools, the face angles are measured and sent to the ROS (node ``face\_angles"). It is vital to notice that angles dealt here are based on rotation vectors shown in Fig.~\ref{newaxes}b. The individual in the frame can pose in any combination of these angles. The reference pose relative to the camera where every angle is zero is shown in Fig.~\ref{cam_pic}\cite{dervinis2006head}.

\section{Head and eye movement imitation}
The primary and most essential part of the project is how it is suitable to estimate head orientation. The first option is selecting a handful of head pose classes and positioning the Agile Eye in those configurations. It is proper to use this approach where the face inside the frame poses the closest to a particular class of poses. Another option could be finding the exact angles in which the face is posing. The last option, which is considered in this paper, is to gather a dataset for pitch and yaw rotations; an then fitting a linear regression model which can be mapped for any range of output that one may desire.

\subsection{Facial landmark}
\begin{figure}[t]
\centerline{\includegraphics[width=.8\columnwidth]{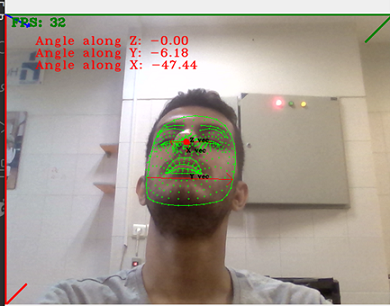}}
\caption{Angles calculated using facial landmarks.}
\label{amir}
\end{figure}

In this study, the Mediapipe library is used. This library contains different solutions such as face detection, human body detection,and tracking, hand tracking, and etc\cite{lugaresi2019mediapipe}. In this project, the 468 landmark solution is used. The Mediapipe face landmark solution is a geometry solution which estimates landmarks such as the face, eye, and mouth in 3 dimensions as shown in Fig.~\ref{amir}. The $x$ and $y$ outputs of the Mediapipe facemech solution are normalized based on the frame size. The $z$ value of a landmark, however, is not the distance to the camera. Every $z$ value of landmarks is relative to each other. For example, if two landmarks have different depths, their $z$ value is measured relative to other $z$ values which may not be useful to calculate the distance to the camera, but it is sufficient to calculate face angles.

\subsubsection{angle along $z$}
In order to measure the angle along the z-axis, as shown in Fig.~\ref{cam_pic}, using a vector in the XY plane seems sound. A vector which starts from the center of the left eye and ends with the center of the right eye is defined. So the slope of this vector gives us a tangent of the angle which leads to calculate the roll angle.

\begin{equation}
m_z = \frac{y_{\text{right eye}} - y_{\text{left eye}}}{x_{\text{right eye}} - x_{\text{left eye}}}
\end{equation}
\begin{equation}
roll: \alpha_z = \arctan(m_z)
\label{zangle}
\end{equation}

\subsubsection{angle along $y$}
for this angle the map of previous vector in XZ plane can be used:

\begin{equation}
m_z = \frac{z_{\text{right eye}} - z_{\text{left eye}}}{x_{\text{right eye}} - x_{\text{left eye}}}
\end{equation}
\begin{equation}
yaw: \alpha_y = \arctan(m_y)
\label{yangle}
\end{equation}

\subsubsection{angle along $x$}
For angle along the $y$-axis a vector for lower nose to the top of the nose is considered:
\begin{equation}
m_x = \frac{y_{\text{top nose}} - y_{\text{lower nose}}}{z_{\text{top nose}} - z_{\text{lower nose}}}
\end{equation}
\begin{equation}
pitch: \alpha_x = \arctan(m_x)
\label{xangle}
\end{equation}

\subsection{phase angle}
Another way to model these angles consists in designing a linear model for them. Between these three angles, the \textbf{roll} angle can be measured with other models such as 68 points, similar to the previous approach, because this angle can only be obtained by getting the position of the eyes. The main problem consists in finding the other two angles because one needs to know the depth of the image to compute them. In this paper, a simple linear regression model is used which a two-dimensional landmark is utilized for obtaining fuzzy in the range between -10 to 10 output. This output can be mapped to any other angle range depending on the robot workspace.

It should be noted that it is the distance of the landmarks with each other which determines the face rotation, not the $(x,y)$ values of each landmark in a frame.

\subsubsection{dataset}
\begin{figure}[htbp]
\centering
\subfloat[Looking center (reference configuration).]{\includegraphics[width=0.32\columnwidth]{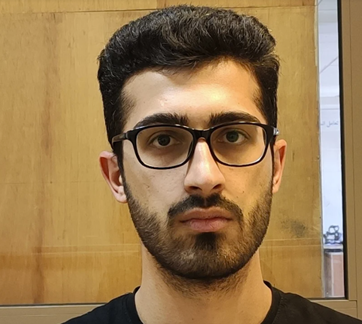}}
\hfil
\subfloat[Looking right.]{\includegraphics[width=0.3\columnwidth]{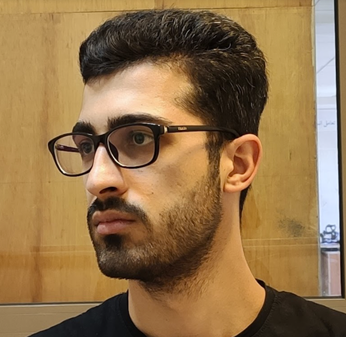}}
\hfil
\subfloat[Looking left.]{\includegraphics[width=0.3\columnwidth]{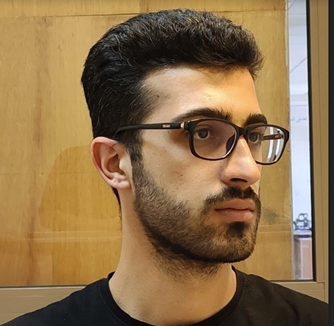}}
\hfil
\subfloat[Looking up.]{\includegraphics[width=0.3\columnwidth]{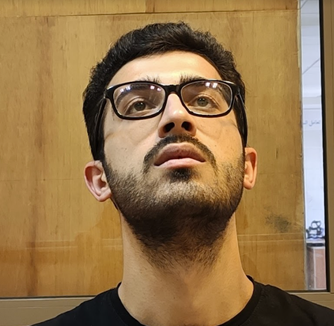}}
\hfil
\subfloat[Looking down.]{\includegraphics[width=0.3\columnwidth]{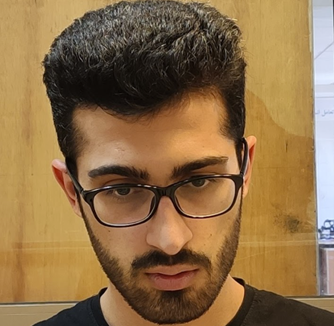}}
\caption{A sample of training data.}
\label{milad}
\end{figure}

The training dataset contains 174 images from different peoplle which are stared in front, right, left, up, and down (Fig.~\ref{milad}). The dataset is mapped to numeric data between -10 and 10. In other words, Fig.~\ref{milad} (d) is mapped to 10 and Fig.~\ref{milad} (e) is mapped to -10 and other poses are mapped in range (-10,10). It should be noted that color background does not affect results and algorithms since the algorithm finds facial poses in any environment.

\subsubsection{preprocessing}
\begin{figure}[t]
\centerline{\includegraphics[width=.6\columnwidth]{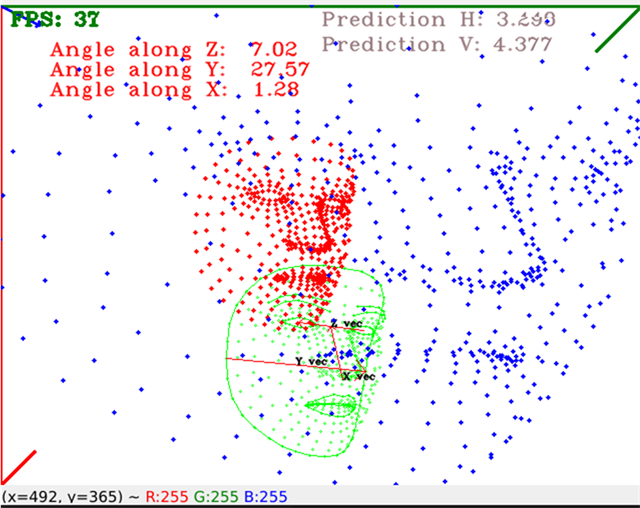}}
\caption{Normalized face mesh is shown with blue dots.}
\label{normal}
\end{figure}

The dataset should be normalized before giving landmarks to the model as shown in Fig.~\ref{normal} because redundant information affects the results, such as distance from the face to the camera or even the size of the face. Also, landmarks extracted from the dataset should be in the center of the frame because the face position in the frame does not affect its angulation.

\subsubsection{training}
A simple linear regression model \cite{MLR} is used to predict face pose. Equation \eqref{linearreg} is the expression of a linear regression model.
\begin{equation}
\begin{multlined}
f(x) = w_0 + w_1x_1 + w_2x_2 + ... = 
\\ w_0 + \sum_{n=1}^{468} w_nx_n
\end{multlined}
\label{linearreg}
\end{equation}
In the above, $x_i$ are the 468 landmark points, and $w_i$ are the weights associated with the points. The network is trained by considering 80\% of the data for training and 20\% for validation. Here, two linear models are used, one for horizontal movement and the other for vertical movement. The model learns the relative distance between the landmarks by adjusting the weights and biases. Nevertheless, this method creates a certain amount of error for combining two vertical and horizontal movements, which is acceptable in the range of the Agile Eye work space which is 15 degrees for the roll, pitch, and yaw angles.


\begin{figure}[t]
\centering
\subfloat[]{\includegraphics[width=.5\columnwidth]{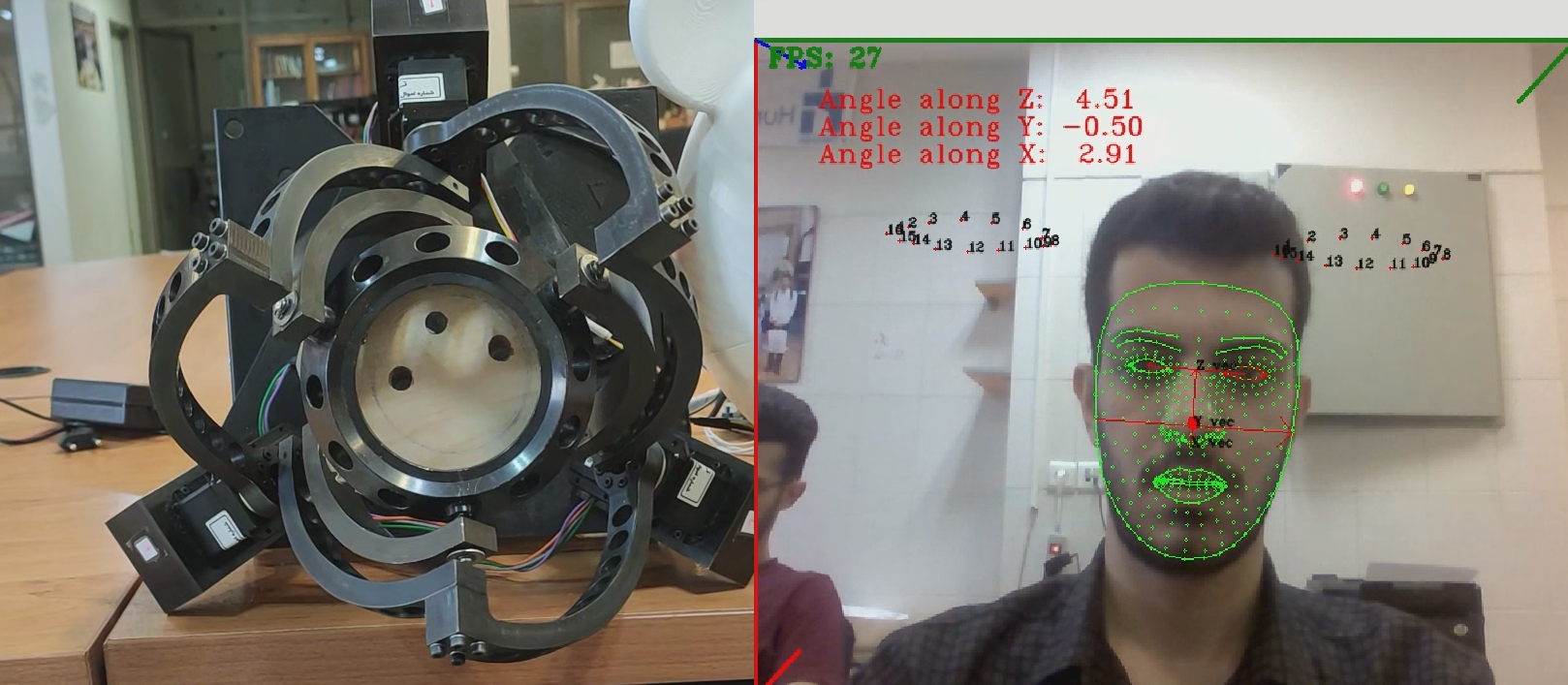}}
\hfil
\subfloat[]{\includegraphics[width=.5\columnwidth]{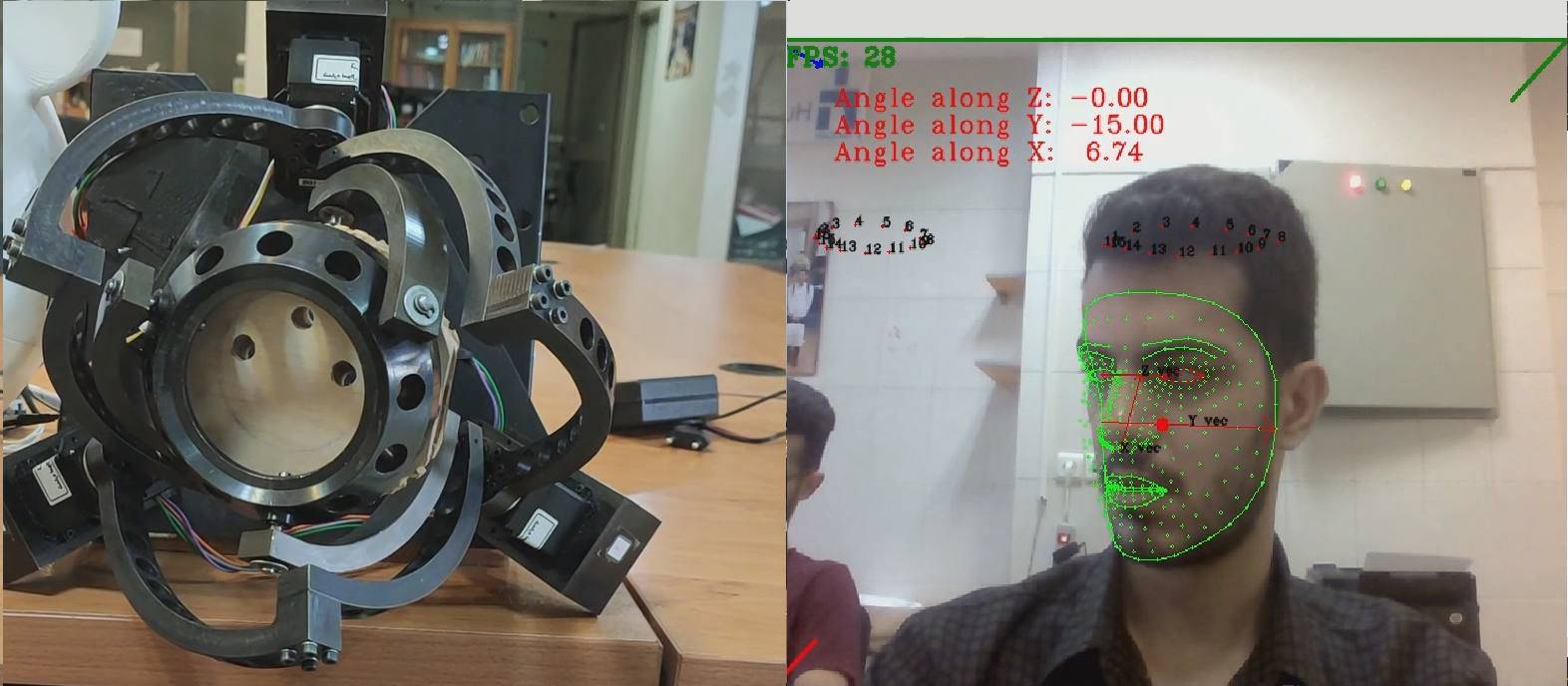}}
\hfil
\subfloat[]{\includegraphics[width=.5\columnwidth]{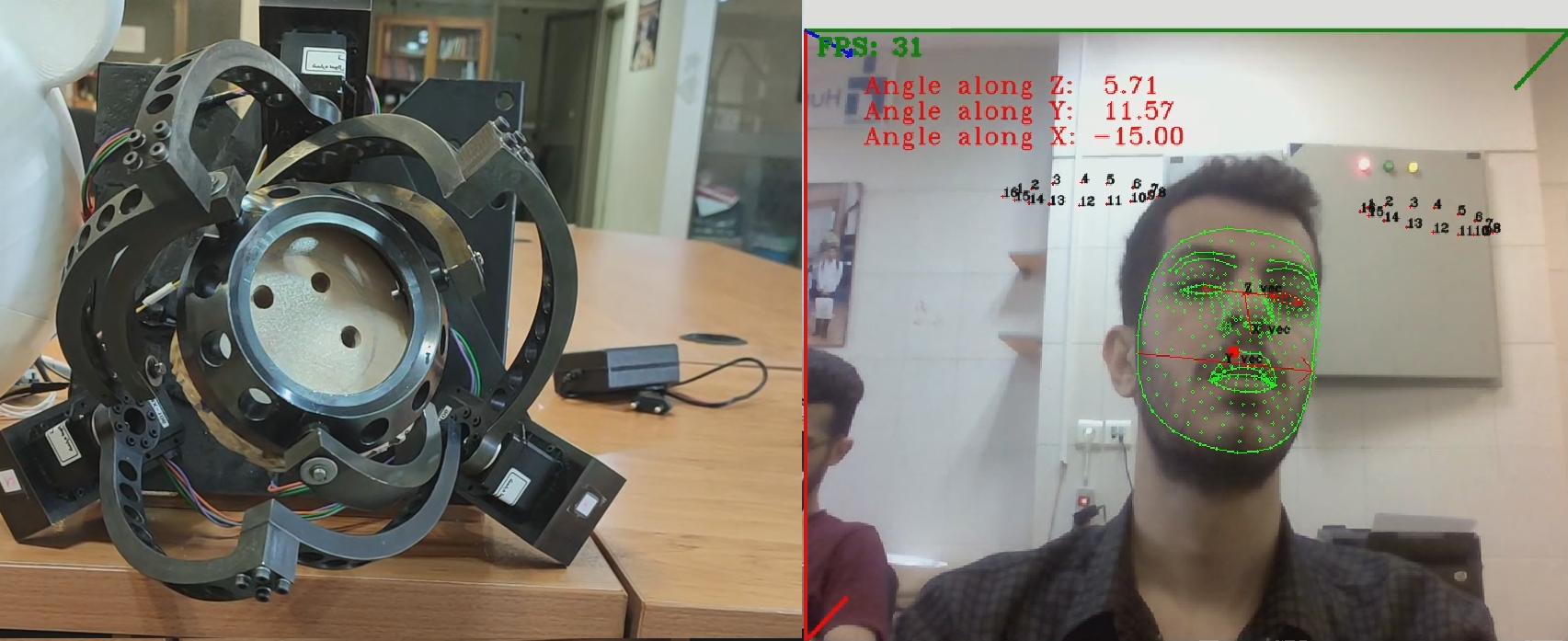}}
\hfil
\subfloat[]{\includegraphics[width=.5\columnwidth]{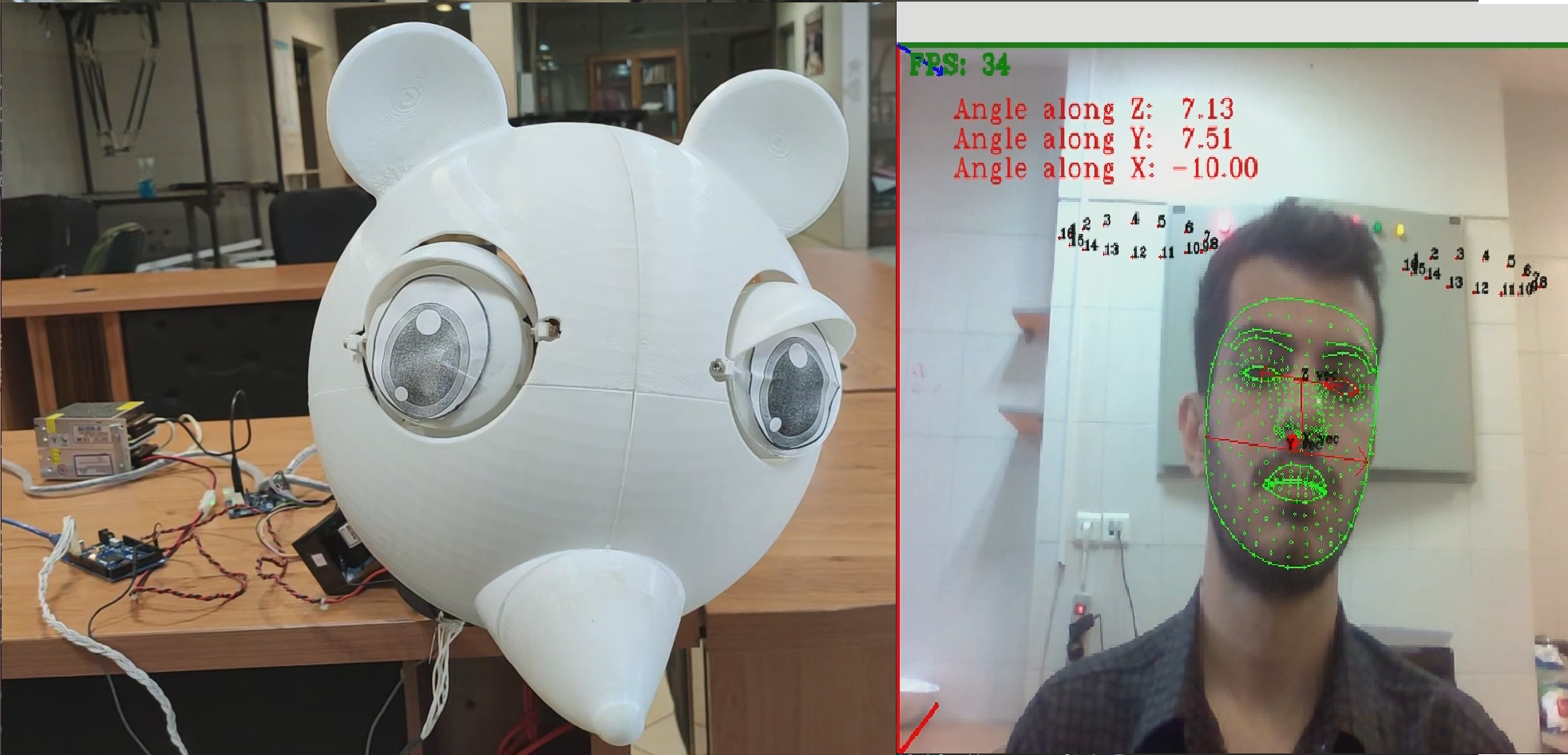}}
\hfil
\subfloat[]{\includegraphics[width=.5\columnwidth]{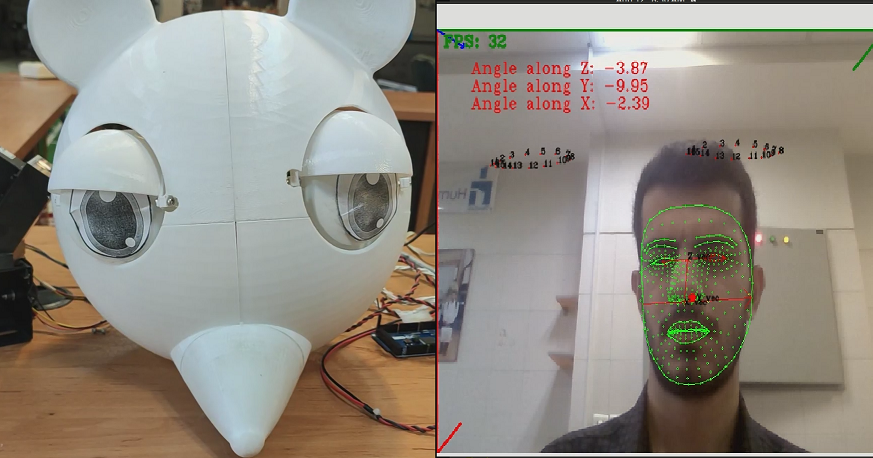}}
\hfil
\subfloat[]{\includegraphics[width=.5\columnwidth]{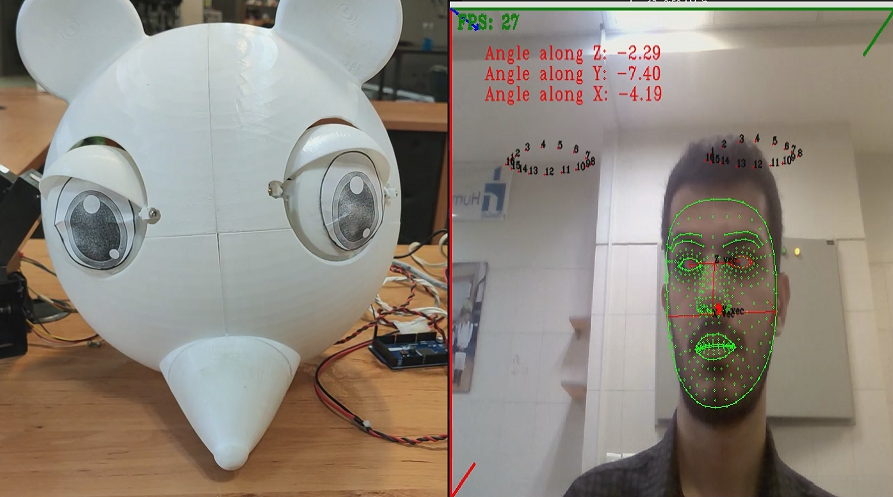}}
\hfil
\subfloat[]{\includegraphics[width=.4\columnwidth]{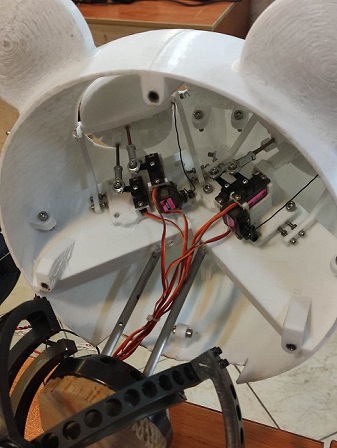}}
\hfil
\caption{Experimental results. (a)-(d) show the different angles taken by the person, while the robot has changed position in the same way.(e) and (f) shows eyelid movements. (g) shows behind the head of the robot and the whole assembly of eye mechanism}
\label{experimentalimages}
\end{figure}

\subsection{Experimental Results}
In this section, as shown in Figure Fig.~\ref{experimentalimages}a-d, by rotating the person's head around the three roll,yaw, and pitch axes, the end-effector of the robot is moved around these three axes, as shown in the mentioned figure, these angles are properly measured and displayed in the frame and applied to the robot.
Although these results shown in Fig.~\ref{experimentalimages}a-d are for only one individual, but for the dataset and other people similar results were obtained and orientations of the end-effector were accurate regardless of who was in front of the camera.
It should be noted that before applying the 3-4-5 motion planning the movement of servos and as a result the rotation of the end-effector of the Agile Eye was shaky and vibrant. This can be destructive to the servos and results would not be acceptable. Especially, it is important that the movements be smooth in real-time at the start and end of a rotation trajectory.
Fig.~\ref{experimentalimages}e,f also shows the blink of a robot based on the calculated area trapped by the eye points inside the frame using the Shoelace formula:

\begin{equation}
\begin{multlined}
area = \frac{1}{2}\left|\sum_{i=1}^{n} det(\begin{bmatrix} x_i  & x_{i+1}\\
									   y_i  &  y_{i+1}\end{bmatrix})\right|
\end{multlined}
\label{linearreg}
\end{equation}

One of the limitations that this system encounters is the speed of the motors. Since the inertia of the head on the robot is high, the speed of whom is posing at different angles should be detected by the software and if it is a high speed, it can be prevented from being applied to the robot.

%

\section{Conclusion}

In this paper, a mechanical setup including a combination of a 3-DOF spherical parallel mechanism was proposed for imitating human head and eye movement. To this end, the Mediapipe library, introduced by Google, was used to find the head's pose accurately. Also, in the linear regression method, a model was trained using the collected dataset and based on this model, the angles of the face were calculated fuzzily. These angles were then applied to the Agile Eye parallel robot  to mimic the movement of the human face. A 2-DOF mechanism was also developed for the robot's eye, which changes rotation based on the person's position in the frame. Moreover, to have a smooth movement while imitating the head movement, a 3-4-5 motion planning strategy was proposed.
Future work intends to develop this platform to interact with children and patients with autism, which can be used in psychological studies for such people.




\end{document}